\documentclass{article}
\usepackage[preprint]{neurips}

\usepackage{microtype}
\usepackage{graphicx}
\usepackage{booktabs} 

\usepackage[utf8]{inputenc} 
\usepackage[T1]{fontenc}    
\usepackage{url}            
\usepackage{amsfonts}       
\usepackage{nicefrac}       
\usepackage{microtype}      
\usepackage{amsmath}
\usepackage{amsthm}
\usepackage{textcomp}
\usepackage{graphicx}
\usepackage{color}
\usepackage{subcaption}
\usepackage{algorithm}
\usepackage{algorithmic}

\newcommand{\eps}{\epsilon}
\newcommand{\EE}{\mathbb{E}}

\usepackage{bbm}

\newcommand{\RR}{\mathbb{R}}

\newcommand{\rv}{\right\Vert}
\newcommand{\lv}{\left\Vert}
 
\newcommand{\ut}{^{(t)}}
\newcommand{\utp}{^{(t+1)}}
\newcommand{\uz}{^{(0)}}  
\newcommand{\tu}{^{(t)}}
\newcommand{\tup}{^{(t+1)}}
\newcommand{\tum}{^{(t-1)}}
\newcommand{\tupp}{^{(t+2)}}

\newcommand{\lp }{\left(}

\newcommand{\rp }{\right)}

\DeclareMathOperator{\erf}{erf}

\DeclareMathOperator{\ReLU}{ReLU}
\DeclareMathOperator{\R}{ReLU}

\global\long\def\a{\rightarrow}

\newtheorem*{rem}{Remark}

\newtheorem{definition}{Definition}
\newtheorem{corollary}{Corollary}

\newtheorem{theorem}{Theorem}
\usepackage{xcolor}
\definecolor{DSgray}{cmyk}{0,0,0,0.7}
\definecolor{DSred}{cmyk}{0,0.7,0,0.7}

\usepackage{hyperref}

\title{Regularization by Misclassification\\ in ReLU Neural Networks}

\author{ Elisabetta Cornacchia, Jan Hązła, Ido Nachum  
    \\ Department of Mathematics
    \\  \'Ecole Polytechnique F\'ed\'erale de Lausanne 
    \\ \texttt{\{elisabetta.cornacchia,jan.hazla,ido.nachum\}@epfl.ch} 
\And
    Amir Yehudayoff
    \\ Department of Mathematics
    \\ Technion - Israel Institute of Technology
    \\ \texttt{ amir.yehudayoff@gmail.com} 
   } 
\begin{document}

\maketitle

\begin{abstract}
  We study the implicit bias of ReLU neural networks trained by a variant of SGD where at each step, the label is changed with probability $p$ to a random label (label smoothing being a close variant of this procedure). Our experiments demonstrate that label noise propels the network to a sparse solution in the following sense: for a typical input, a small fraction of neurons are active, and the firing pattern of the hidden layers is sparser. In fact, for some instances, an appropriate amount of label noise does not only sparsify the network but further reduces the test error. We then turn to the theoretical analysis of such sparsification mechanisms, focusing on the extremal case of $p=1$.  We show that in this case, the network withers as anticipated from experiments, but surprisingly, in different ways that depend on the learning rate and the presence of bias, with either weights vanishing or neurons ceasing to fire. 
\end{abstract}








\section{Introduction}

Neural networks generalize well while being very expressive,  with a training process that produces solutions with zero training error. It is thus expected that some form of {\em implicit regularization} by the training algorithm makes neural networks extremely useful.

Our focus is on the role played by label noise in such implicit regularization. We study scenarios where at each iteration during  training, the true label is changed with probability $p$ to a uniformly random label. With $p=0$ there is no noise, and with $p=1$ the label is uniform.

{\em Why would it make sense to add label noise on carefully collected data?}
The answer is subtle.

Noise is often thought of as a major cause of difficulty in learning. 
But noise can sometimes improve the accuracy of the model; a canonical example is label smoothing~\citep{LS}.

Some recent works~\citep{blanc2019implicit,LSH} explore when and how label noise affects neural networks. 
Our main finding is that training with label noise results in \emph{sparser activation patterns}. 
Namely, neurons fire less often in networks trained with label noise.
Here is the measure of  sparsity
that we consider in this paper.

\begin{definition}\label{def:num_act}
Let $N(x)=W_2\cdot \ReLU (W_1 \cdot x+B_1)+B_2$ be a fully connected network with one hidden layer. For $x$ in the dataset, the number of active neurons is $A_N(x) = |\{i~|~w_i \cdot x+b_i > 0   \}|$  where $(w_i,b_i)$ correspond to the weights of neuron $i$ in the hidden layer.
The typical number of active neurons is $\EE_x A_N(x)$,
where $x$ is uniformly distributed in the dataset. \end{definition}

To see how  label noise sparsifies the firing pattern of a network, we  start from a broad question. How does the network change when it is presented with a mislabeled sample?  One natural approach for answering this question is to study the evolution of norms in the network during training. In some applications, norms are explicitly regularized, and evidence suggests that  regularization helps the network to generalize better (see Chapter 7 in~\cite{dlbook}). 

\subsection*{Our Contribution}

As a first theoretical observation, in Section~\ref{sec:norm}
we prove that when a general neural network 
is presented with a mislabeled sample, the Frobenius norm of its weight matrices decreases (see Theorem~\ref{thm:w_decay}). This is consistent with the fact that sometimes explicit $L2$ regularization is not required in practice to achieve  comparable generalization, as observed in~\cite{van2017l2}.

The decrease of norms suggests the possibility that if the network is given a ``too hard'' dataset then the weights are nullified and the network effectively dies.  

Indeed, in Section~\ref{sec:theory} we present a theoretical analysis
of the extreme setting of completely noisy data ($p=1$).
We identify several different mechanisms of network decay in that case, 
which are presented in Theorems~\ref{thm:v_van},~\ref{thm:network_relu_nobias},~and~\ref{thm:step_size}. 

 Theorem~\ref{thm:v_van} proves that under pure label noise ($p=1)$ over the normal distribution $\mathcal{D}=\mathcal{N}(0,I_d) $, the weights of a single neuron will roughly decay to zero. 
 
 We then move to studying networks with multiple
 neurons arranged in one hidden layer.
 Empirically, 
 we show that 
 they decay as well and we identify two modes. Without bias neurons,  all the weights roughly decay to zero. With bias neurons, the weights do not decay to zero, but all ReLU neurons die during training, so after training, they do not respond to any input. 
 
 Theorem~\ref{thm:network_relu_nobias} proves that after training under pure label noise over the distribution $\mathcal{D}=U\{e_1,...,e_d\}$ (uniform over the standard basis), a network without bias neurons roughly outputs zero for all inputs. Theorem~\ref{thm:step_size} shows that in this setting there are two different modes of decay that depend on the size of the learning rate. For a large learning rate, all the ReLU neurons do not respond to any input; this shows that bias neurons are not always the cause of ReLU death.  For a small learning rate, the output neuron dies, but a sizable fraction of neurons remains active. We observed this behavior also empirically; larger learning rates induced greater sparsity.

Extreme label noise causes unfruitful results, as one would expect.
What about smaller amounts of label noise? In Section~\ref{sec:experiments}, we empirically observe that training with label noise can be beneficial.  Figure~\ref{fig:square}  presents an example (see Section~\ref{sec:cube}) where the role of label noise is critical for learning.  Without label noise, the network easily fits the data but generalizes poorly with a test error of $49\%$. Modiying $p$, from $0$  to $0.2$, makes a tremendous difference and the test error drops to $21\%$. 

Over the training set, for a one hidden layer network with 240 neurons,  the typical number of active neurons with label noise is roughly $47$ (i.e. $20\%$) whereas without label noise it is roughly $94$ (i.e. $40\%$). These quantities remain the same even when measured over the test set. This may give an insight on why training with label noise yields better
test accuracy. The increased sparsity comes at a cost: the training time is longer.



\begin{figure}[h]
\begin{center}
 \begin{subfigure}{0.45\textwidth}
\includegraphics[width=0.95\linewidth]{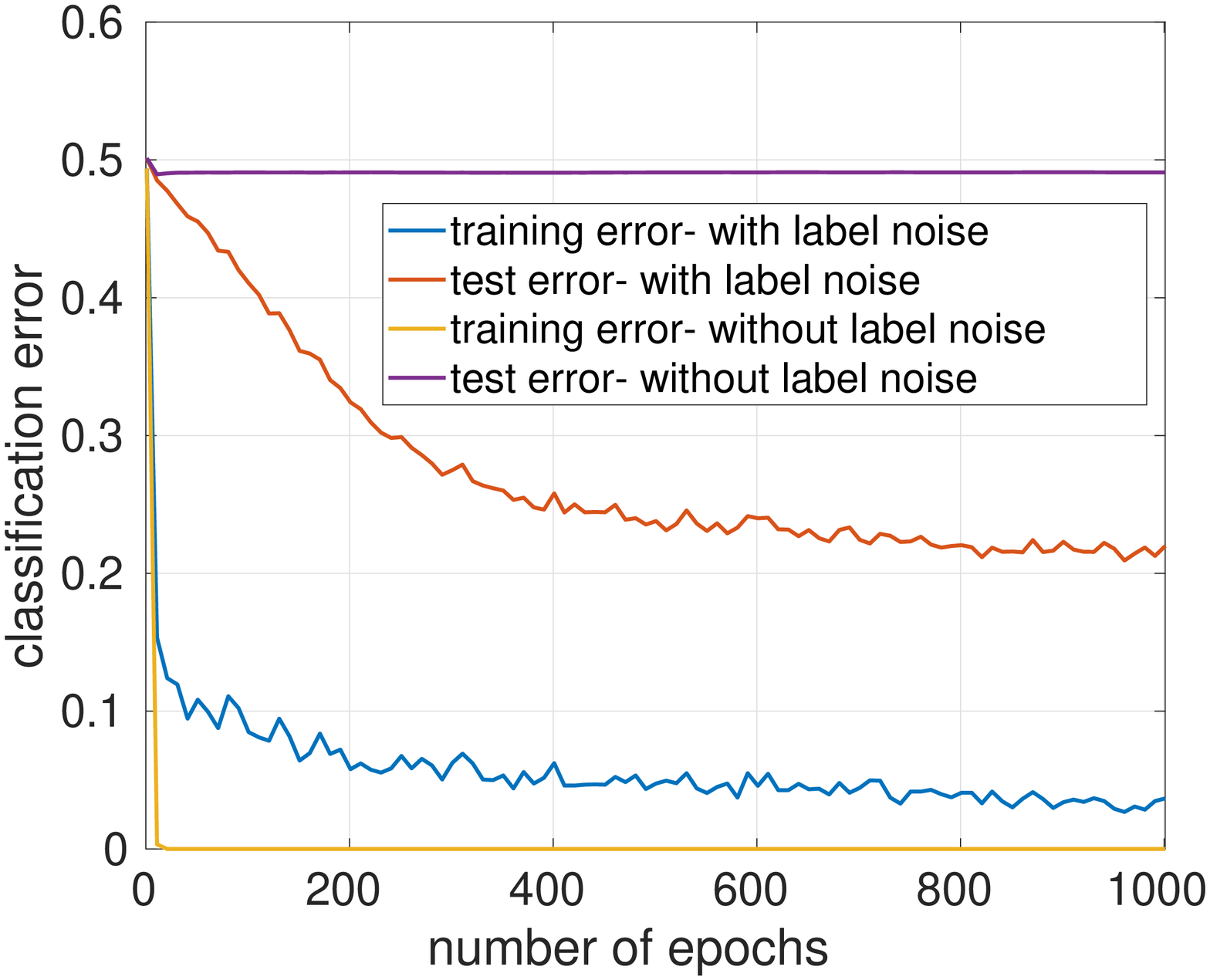} 
\caption{error evolution}
\label{fig:sq_error}
\end{subfigure}\hfill
\begin{subfigure}{0.45\textwidth}
\includegraphics[width=0.95\linewidth]{ 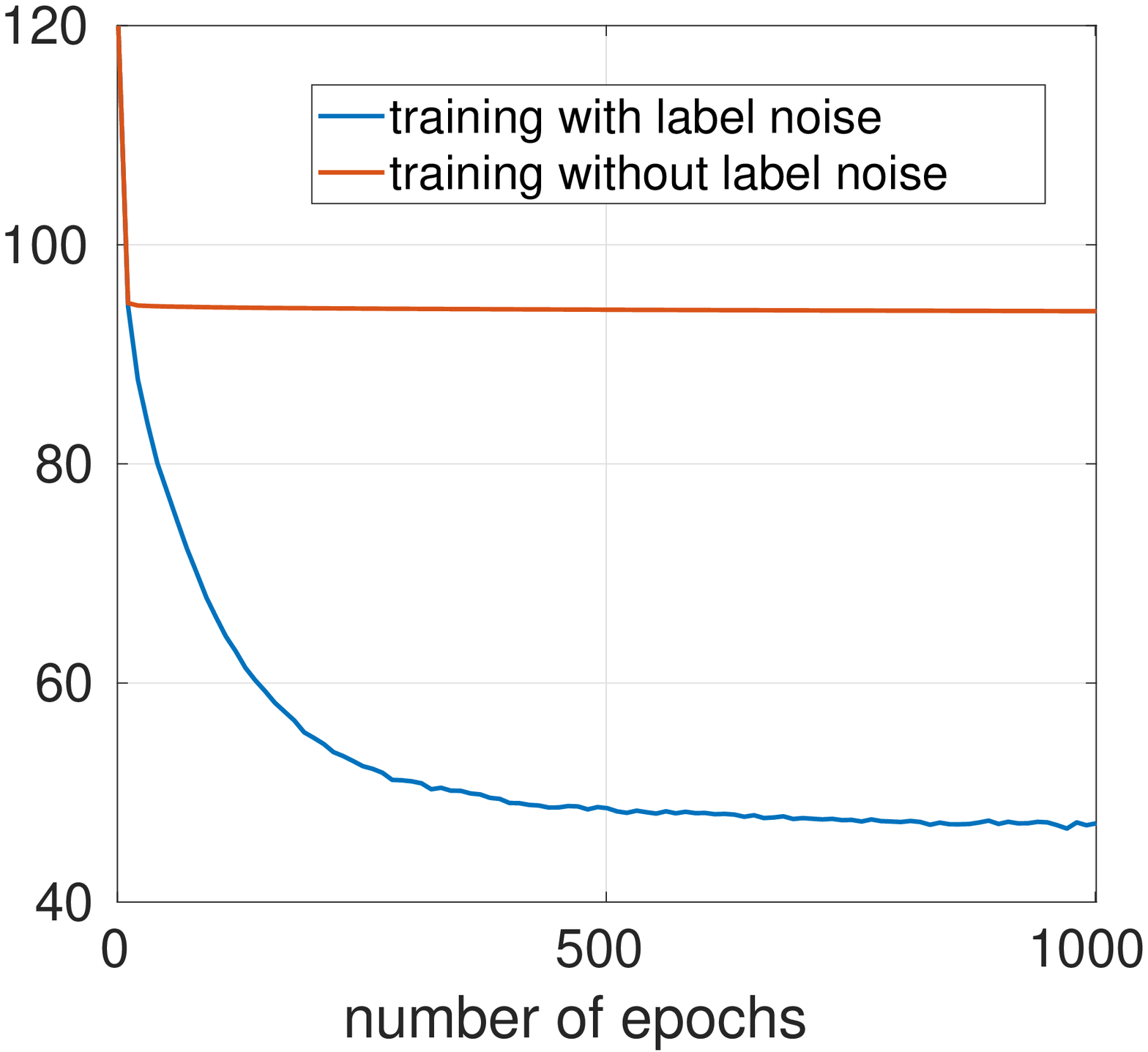}
\caption{typical number of active neurons}
\label{fig:sq_active}
\end{subfigure}
 \caption{ Learning the hypercube function.  The input dimension is $d=60$,
 there are $4d$ neurons in the hidden layer 
 and the learning rate is $h=1/d$. The plots were averaged over $20$ independent~runs.}
\end{center}
\end{figure}\label{fig:square}

The intermediate value theorem suggests an additional perspective. Consider binary classification with a label noise parameter $p \in [0,1]$.
When $p=0$ there is no label noise,
and when $p=1$ the label is completely random.
For every $p$, consider the value $a(p)$ equal
to the typical number of active neurons with noise parameter $p$.
The above discussion suggests that $a(0)$ is much larger than $a(1)$.
Tuning $p$ from $0$ to $1$
allows to tune the sparsity from $a(0)$ to $a(1)$ (a rigorous analysis of the continuity of $a(p)$ is in the  appendix).
And possibly, there  exists a value of $p$ where the network still fits the data but with low sparsity. 
This, in turn, may improve generalization, as we examine further in the experiments in Section~\ref{sec:experiments}.

\section{Related work}

Even though overparametrized neural networks are well known to perfectly fit 
random examples, 
there is a line of work arguing that adding independent label noise to
every iteration of the SGD effectively results in a problem that cannot
be perfectly fit, possibly improving generalization and resulting in
smoother solutions. Some, mostly empirical, works in this vein
include~\cite{Hanson1990ASV,Clay1992FaultTT,Murray1994EnhancedMP,an1996effects,breiman2000randomizing,Rifai2011AddingNT,sukhbaatar2014learning, maennel2020neural}.
Our contribution is pointing out the connection of this behavior to
the regularizing property described above via sparsity.

A good representative is a recent work~\citep{blanc2019implicit} that investigates label noise of one-dimensional 
regression. In their setting, they add independent noise to labels in the training set, which results in data that cannot be perfectly fit. They then argue that this steers the gradient descent towards solutions where the gradient of an implicit regularizer vanishes in certain directions.
Our empirical results show a similar behavior for classification problems.
Interestingly,~\cite{blanc2019implicit} argue that this ``noise regularization''
does not occur in networks with only one layer of trainable weights. However, our experiments for MNIST yield similar
results even if only one layer of weights is trained.


A theoretical reason to study learning pure label noise  
is also given by~\cite{Abbe2020PolytimeUA}. This paper studies the ``junk flow'',
a notion that acts as a surrogate for the number of queries in lower-bound techniques for gradient descent. 
In particular, understanding the dynamics of learning under noise can explain why a randomly initialized network will fail at learning functions like large parities. Such functions might produce data that appear to the network as random data, and the ReLU neurons will die before the network gets to be trained. 

Dying ReLUs are well observed in practice.  \cite{Lu2019DyingRA} study this phenomenon from an initialization perspective and suggest alternative initialization schemes. \cite{Arnekvist2020TheEO, Douglas2018WhyRU} treat this problem, as in our case, from the weight dynamics perspective. ~\cite{sparse1} study a related notion, filter level sparsity. They conduct an extensive empirical study on the mechanisms that allow sparsity to emerge.

Implicit regularization of neural networks has been observed and studied in numerous works such as
\cite{
Du2018AlgorithmicRI,hanin2018neural,Neyshabur2015InSO,Gunasekar2018ImplicitBO,Soudry2018TheIB,soudry2017implicit}. For instance, \cite{soudry2017implicit} shows that for monotonically decreasing loss functions, linear predictors on separable data converge to the max-margin solution. 

\section{Preliminaries}
\label{sec:prel}

A neural network consists of a consecutive application of an affine transformation followed by a non-linearity~$\sigma$:
 \nopagebreak \begin{equation}\label{eq:net_withbias} 
    N(x)=   W_k\cdot\sigma(...\sigma(W_2\cdot \sigma(W_1 \cdot x+B_1)+B_2)...)+B_k
\end{equation}
\nopagebreak We typically work with the standard $\ReLU(x)=\max\{0,x\}$ non-linearity. 
More generally, we consider homogeneous non-linearities.
A function $\sigma:\RR \a \RR $ is homogeneous if it is piecewise differentiable and if for all $x\in \RR \backslash \mathcal{K}$ it holds that $f'(x)\cdot x=f(x)$, where $\mathcal{K}$ is a finite set.

We focus our attention on SGD with mini-batch of size one and label noise parameter $p\in [0,1]$. In each iteration, a single sample $x$ is chosen uniformly at random from the dataset and its label $y$ is changed with probability $p$ to a uniform random label.

We consider samples with $x\in \RR ^d$ and $y\in Y$, where $Y$ is a finite subset of $\mathbb{R}$. The weights are then updated according to the gradient of a loss function $\mathbb{L}(\textbf{W},\textbf{B},x,y)$, where $\textbf{W}=(W_1,...,W_k)$ and $\textbf{B}=(B_1,...,B_k)$. For a learning rate $h>0$, the update rule of the weights is
$$(\textbf{W}^{(t+1)},\textbf{B}^{(t+1)})=(\textbf{W}^{(t)},\textbf{B}^{(t)})-h \nabla_{(\textbf{W}^{(t)},\textbf{B}^{(t)} )} \mathbb{L}(\textbf{W}^{(t)},\textbf{B}^{(t)},x,y)$$
The update rule above dictates how the weight matrices and the bias vectors of the neural network evolve with time
(the weights are possibly shared as in convolutional neural networks). 

For the theoretical analysis, we consider
binary classification with
$Y=\{\pm 1\}$ and
{\em 0-1 surrogate} loss functions. That is,
loss functions of the form 
$\mathbb{L}(\textbf{W},\textbf{B},x,y)
= L(-y N(x))$, where $L$ is
increasing, convex, piecewise differentiable and its derivative at zero
(from the right) is positive.
Two examples are the hinge loss with parameter $\beta \geq 0$ defined by $L(\xi)=\R(\beta+\xi)$ 
and the logistic loss defined by $L(\xi)=\log ( 1+\exp(\xi))$.

Finally, we remark that training with label noise 
is closely related to label smoothing, introduced in~\cite{LS}. The purpose of label smoothing is to prevent the network from being overconfident in its predictions. In this framework, the activations in the output layer $\{z_y=N_y(x)\}_{y \in Y}$ are treated as a probability distribution $p_y = \frac{ \exp(z_y) }{\sum_{y'}  \exp(z_{y'}) }$. Then, for example, assuming the correct class is ``1'', this distribution is optimized against the distribution $\tilde p=\left( 1-p+p/|Y|,p/|Y|,...,p/|Y|\right)$, instead of $(1,0,...,0)$, using the cross-entropy loss $-\sum_{y \in Y} \tilde p_y \log p_y$. 

The parameter $p$ of label smoothing is the equivalent quantity for the parameter $p$ of learning with label noise. Label smoothing  distributes a probability mass $p$ uniformly across all labels and label noise assigns all of the probability mass to a random label with probability $p$. So in some sense,  label smoothing is label noise applied in expectation.   

\section{Implicit Norm Regularization}
\label{sec:norm}

The following theorem shows that, when using a small enough learning rate,
the Frobenius norm $\lv \cdot \rv \equiv \lv \cdot \rv _F$ of the weights decreases when presented with a mislabeled sample.
This holds, e.g., for the hinge and logistic loss, as well as for activations such as ReLU or Leaky ReLU.

For simplicity, we consider binary classification 
$y\in\{\pm 1\}$ and for this theorem to hold, we consider neural networks of the form  \begin{equation}  \label{eq:net_nobias}
    N(x)=   W_k\cdot\sigma(...\sigma(W_2 \cdot \sigma(W_1 \cdot \Tilde{x}))...),
\end{equation}

where $\Tilde{x}=(x,1)$. The bias terms implicitly appear in the matrices $W_i$,
so such networks can express the same functions as the networks presented in equation~\eqref{eq:net_withbias}. However, the  corresponding gradients will be different. Nevertheless, the weight decay (possibly flatlining at a constant
level rather than decaying all the way to zero)
still holds empirically for networks of type~\eqref{eq:net_withbias} (for example, see Figure~\ref{fig:bias_weight_decay}). 

\begin{theorem} \label{thm:w_decay}
Let $N\tu:\RR^n \a \RR$ be a neural network at time $t$ of training, of any architecture (including weight sharing), with homogeneous activation functions (not necessarily the same for each neuron), a 0-1 surrogate loss function $L$, and let $(x,y)$ be a sample such that $y N\tu(x)<0$. There exists $h_0>0$ such that for all $0<h<h_0$ and any layer
of weights $W$  of $N\tu$ it holds that $\lv W\tup  \rv ^2 <  \lv W\tu  \rv ^2 $. 
\end{theorem}

Theorem~\ref{thm:w_decay} is derived by applying a 0-1 surrogate loss to Corollary~2.1 and Theorem~2.3 in~\cite{Du2018AlgorithmicRI} (we provide a proof in the appendix, including self-contained simpler versions).


Considering the hinge loss with $\beta=0$, updates  only occur for misclassified data points. In this case, Theorem \ref{thm:w_decay} immediately implies a special corollary for gradient flow training:
$ \dot{\textbf{W}}= -\nabla _ { \textbf{W}  } \EE_{x,y}\ \mathbb{L} $.

\begin{corollary} \label{w_mon}
Let the setting be as in Theorem~\ref{thm:w_decay},
but with hinge loss with $\beta=0$ and gradient flow
training. Then,
$\lv \textbf{W}\tu  \rv$ is monotonically decreasing as a function of $t$. 
\end{corollary}

\section{Pure Label Noise Leads to  Dead Networks}
\label{sec:theory}

Corollary \ref{w_mon} summons a natural theoretical question: is it possible for the weights to decay to zero,
effectively cutting off all connections in the network?

Or stated more generally,
\emph{what are the underlying mechanisms that cause a network to die during training? }

Theorem \ref{thm:w_decay} suggests an answer.  As long as the network is presented with enough misclassified data points, there is a
pressure to decrease the weights. An extreme example is learning a random function, or almost equivalently, presenting the network with random labels.
Since this case is theoretically tractable, in this section we study it in more detail.

\begin{definition}
A neural network  is trained under pure label noise if at each step $(x,y)$ 
is chosen independently according to $x \sim \mathcal{D}$ and $y\sim U\{ \pm 1\}$ for some
distribution $\mathcal{D}$.
\end{definition}

We use the above definition to answer this question in several examples that  model some of the dynamics  of a network on different scales, from a single neuron to a larger network.

We start with a single neuron. For many typical cases, the following theorem shows that from any initialization, the norm of the weight vector  decreases to roughly zero.

\begin{theorem} \label{thm:v_van}
 Assume the loss function satisfies $\sup_{x} L'(x)=M$ and (a) $\lim_{x\rightarrow 0^+} L'(x) - \lim_{x\rightarrow 0^-}  L'(x) = m > 0 $  (as in hinge loss) or (b) $L''(0)=m>0$  (as in cross entropy)   and let  $V^{(0)}\in \RR^{d}$ be  an initialization for a single neuron.  With high probability, as $h$ goes to zero with other parameters fixed,
 after training under pure label noise with $\mathcal{D}= \mathcal{N}(0,I_d)$, and after at most 
 
 (a) $T=O(\frac{\|V^{(0)}\|}{mh} )$ steps,  $\| V^{(t)}\|$  decreases to 
$O\Big(dh \max\{\frac{M^2}{m}, 1 )\}\Big)$. 
 
(b) $T=O(\frac{\|V^{(0)}\|}{mdh^{3/2}} )$ steps,  $ \|V^{(t)}\|$  decreases to 
$O\Big( d\sqrt{h}\max\{\frac{M^2}{m}, 1 \}\Big)$.

\end{theorem}

\paragraph{Proof in expectation over $\mathcal{D}$.}
For ease of presentation, the full version of Theorem~\ref{thm:v_van} does not include a bias term. 
 So here, we prove a weaker statement that the norm decreases in expectation for the hinge loss with $\beta =0$ \emph{with a bias term}. 
 
 Let $W\tu=(V\tu, b\tu)$ where $V\tu\in \RR^d$ and $b \in \RR$ and $\tilde{x}=(x,1)$.
  At each step, the norm of $W \ut$ changes according to $\lv W \utp \rv ^2=(W\ut +hy\tilde{x} ) \cdot (W \ut +hy\tilde{x})  = \lv W\ut \rv ^2 +2yW\ut \cdot \tilde{x} h +\lv \tilde{x}\rv ^2 h^2$ if  
  $yW\ut \cdot \tilde{x}<0$. Otherwise, there is no change.
Notice that when there is a change, we can take a small enough learning rate $h$, so the norm will  decrease as the leading term is  $2yW \ut \cdot \tilde{x}  h$, which is negative by definition.

By the rotational symmetry of $x$, 
we have $W\tu \cdot \tilde{x}\sim \mathcal{N}(b\tu,\lv V\tu\rv^2)$. 
We now show that, conditioned on  $W\tu$, the expected decay rate of $ \lv W\ut \rv^2$, $$r\tu=\EE_{x,y} [y(W\tu\cdot \tilde{x})\textbf{1}_{y(W\tu\cdot \tilde{x})<0}],$$   is bounded by
$r\tu < - C \cdot \lv W\tu\rv $ for some universal constant $C>0$. 

To see why this holds, let $\sigma=\lv V\tu\rv$ and $\mu =b\tu$. Assume w.l.o.g. $\mu\geq 0$, then,  $r\tu = -\frac{1}{2\sqrt{2\pi}\sigma } \int_0^\infty x\exp(\frac{ -(x-\mu)^2}{2\sigma^2} )  +x\exp(\frac{ -(x+\mu)^2}{2\sigma^2} ) dx =-\frac{\sigma}{\sqrt{2\pi}} \exp (\frac{ -\mu^2}{2\sigma^2} ) - \frac{\mu}{2} \erf (\frac{ \mu}{\sqrt{2}\sigma} ) $. 

If $\mu/\sigma \leq 1$, $r\tu < -\frac{\sigma}{\sqrt{2\pi}} \exp (\frac{ -\mu^2}{2\sigma^2})<  -\frac{\sqrt{\sigma^2+\mu^2}}{2\sqrt{2\pi}} \exp  (-\frac{ 1}{2})< -c_1 \cdot \lv W\tu\rv$.

If $\mu/\sigma \geq 1$,  $r\tu < - \frac{\mu}{2} \erf (\frac{ \mu}{\sqrt{2}\sigma} ) < - \frac{\sqrt{\sigma^2+\mu^2}}{4} 
\erf (\frac{ 1}{\sqrt{2}} ) < -c_2\cdot \lv W\tu\rv$.

We pick $C=\min\{c_1,c_2\}$. This means that as long as  $\lv W\tu \rv > dh/C $,
\begin{align*}
&\left(\EE\lv W\tup\rv\right)^2 -\lv W\tu \rv ^2\leq \EE \lv W \tup \rv^2-\lv W \tu \rv^2
\\&\quad= 2h r\tu + h^2 \EE \lv x\rv^2
< -Ch \lv W\tu\rv
\end{align*}

and therefore
\begin{align*}
&\EE \lv W \tup \rv- \lv W \tu \rv
=\frac{\left(\EE\lv W\tup \rv\right)^2- \lv W\tu \rv^2}
{\EE \lv W \tup \rv+ \lv W \tu \rv}
\\&\quad
<\frac{-Ch\lv W\tu\rv}{2\lv W \tu \rv}
=-Ch/2\;.
\end{align*}
Note that this expression predicts that the norm will roughly decrease at a linear rate. This is 
empirically verified in Figure~\ref{fig:single}.   \hfill\qedsymbol

\begin{rem}
Although the expectation in the proof above is bounded away from zero, some updates during SGD will inevitably increase the norm. In the
full proof in the appendix,
we use a general 0-1 surrogate loss and concentration bounds
to show that nevertheless most of the time the norm
decreases.

\end{rem}

Will the  theorem above  hold for  neural networks with one or more hidden layers? 
It might come as a surprise that it depends on whether  the network has bias neurons or not. For a network with one hidden layer, ReLU activations, and the cross entropy loss, we provide empirical evidence for
training under pure label noise over the normal distribution.

\textbf{Without bias neurons.}  The network weights decay to zero (see Figure \ref{fig:weight_decay}). In this case, it is important to note that the neurons cannot stop firing for all inputs. In fact, because $\mathcal{D}= \mathcal{N}(0,I_d)$ is symmetric, if a neuron does not fire for $x$, it will fire for $-x$.


\textbf{With bias neurons.}  The network weights do not decay to zero.  In this case, the network experiences another kind of decay, where
all the ReLU neurons stop firing for almost all inputs from the distribution $\mathcal{D}= \mathcal{N}(0,I_d)$ due to significant negative
drift in the bias weights.
We call this kind of decay ``ReLU death''. Figure \ref{fig:bias_decay} provides an explanation. 
On the one hand, close to initialization, when the bias
terms do not dominate other weights,
the norm of the weights of the hidden layer decreases
by a mechanism similar   to Theorem~\ref{thm:w_decay},
see Figure~\ref{fig:bias_weight_decay}. 
This means that a typical preactivation value without the bias term decreases.
On the other hand, the bias term of a neuron is monotonically decreasing as in Figure~\ref{fig:bias_evo} (at initialization, this
can be seen from backpropagation formulas). Eventually, the bias becomes so negative that it prevents any input from activating the neuron.

To summarize, a neural network trained under pure label noise over the normal distribution
experiences  neural ``death'': all the weights decay to zero or all the neurons in the hidden layer do not respond to any input. However, the exact
mode of this process can be different depending on the
architecture, eg., the presence of bias neurons as well as 
the data distribution. 

In the following two theorems (proofs in the appendix) we consider a simpler data distribution  (the uniform distribution over the standard basis) and networks with one hidden layer and no bias neurons. We follow the theoretical setting appearing in~\cite{SGD} where we train only the hidden layer and only considering the hinge loss. This allows for a clear presentation of the underlying mechanisms behind the behavior of SGD under pure label noise. 

\begin{theorem}\label{thm:network_relu_nobias}
Let $N$ be a network with one hidden layer with $2k$ ReLU
neurons: $N(x)=V\cdot ReLU(Wx)$ and no bias neurons.
Assume only $W$ is trained and that we fix $V\in\mathbb{R}^{2k}$ to be
split equally between $1$ and $-1$:
$V=(1,\ldots,1,-1,\ldots,-1)$.

Then, for any initialization of $W$, if $N$ is trained with the hinge loss ($\beta=0$)  under pure label 
noise with $\mathcal{D}=U\{e_1,\ldots,e_d\}$
(uniform over the standard basis), after a finite number
of steps we will have $|N(x)|<2kh$ for every $x\in\mathcal{D}$.
\end{theorem}

Theorem~\ref{thm:network_relu_nobias} shows that
the output of the network becomes roughly zero for every input $\{e_1,\ldots,e_d\}$.
The next theorem shows that the type
of neural death depends on  the learning
rate size (even without existence of bias neurons).
If the learning rate is small enough, the top neuron dies, its output becoming roughly zero. If the learning rate is large enough, all neurons in the hidden layer die.

\begin{theorem}\label{thm:step_size}
Let $N$ be a network with one hidden layer with $2k$ ReLU
neurons: $N(x)=V\cdot ReLU(Wx)$ and no bias neurons.
Assume only $W$ is trained and that its weights are initialized iid $U[-1,1]$  and that we fix $V$ to be
split equally between $1$ and $-1$:
$V=(1,\ldots,1,-1,\ldots,-1)$.

As $k$ grows, w.h.p., if $N$ is trained with the hinge loss ($\beta=0$)
under pure label
noise with $\mathcal{D}=U\{e_1,\ldots,e_d\}$
(uniform over the standard basis), we will have after training:
\begin{enumerate}
    \item For a learning rate  $h\geq1$, for all $i$ it holds $\ReLU (We_i)=\overrightarrow{0}$.
    \item For a learning rate  $h\leq1/k$, for all $i$ it holds $\ReLU (We_i)\neq \overrightarrow{0}$. Furthermore, at most $k+o(k)$ (i.e.,
    $1/2+o(1)$ fraction)
    of the coordinates of the vector $\ReLU(We_i)$ are zero. 
\end{enumerate}
\end{theorem}

The motivation for Theorem \ref{thm:step_size} is to show an example
on a slightly larger scale, where even without bias neurons, the ReLU neurons can stop firing without all the weights decaying to zero.
Such behavior can be observed when training over datasets like MNIST
and pictures in general, where the data consists of vectors with non-negative entries (with analogy to the standard basis). However, for MNIST, the learning rate in which all the neurons die is much smaller than the rate $h\geq 1$ that appears in Theorem~\ref{thm:step_size}. For example, empirically,
for MNIST item $1$ holds for the rate $h=0.001$.

\section{Noise Induces Sparsity --- Experiments}
\label{sec:experiments}
The discussion above suggests  that if the data  given to a ReLU  network is very noisy, it might ``destroy'' the network during training.  We now verify this empirically. At the same time, we  show instances where the network benefits from adding some noise to the data.

In this section, we experiment and progress from exactly following our theoretical setting (a single neuron, binary classification, pure label noise, mini-batch size $1$) to a real-life neural network (ResNet20, CIFAR-10,  label smoothing,  mini-batch size $128$). Our main finding, of sparse activation patterns, does not depend on the exact noise model. So accordingly, we do not focus our discussion
on the differences between them.




For a clear and consistent presentation, in all experiments (unless specified differently), we  use input of dimension $d=30$, ReLU activations, the  binary cross entropy loss,  SGD with  a mini-batch of size one, a fixed learning rate
 $h=1/d^2$, and  a uniform i.i.d.~weight initialization $\sim U[-\sqrt{3},\sqrt{3}]$ ($\sqrt{3}$ for unit variance).
 To the best of our knowledge varying any of these parameters
 does not affect our main findings. For example, increasing the size of SGD batch
 results in similar outcomes.

\subsection{A single neuron under pure label noise}

\begin{figure}[ht]
\vskip 0.1in
\begin{center}

\includegraphics[width=0.5\linewidth]{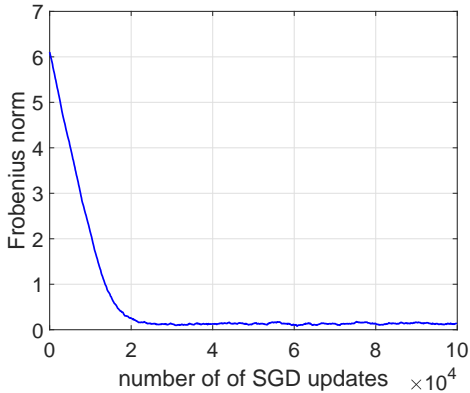} 
\caption{Norm evolution --- learning noise with $x\sim N(0,I_d)$. }
\label{fig:single}
\end{center}
\vskip -0.1in
\end{figure}

We start with a single neuron. Figure \ref{fig:single} shows that when a neuron is trained under pure label noise, the norm of the weights decays at a linear rate up to an equilibrium point of norm $dh$, as predicted by Theorem~\ref{thm:v_van}.

\subsection{Network decay under pure noise}
\begin{figure}[ht]

\begin{center}
\includegraphics[width=0.5\linewidth]{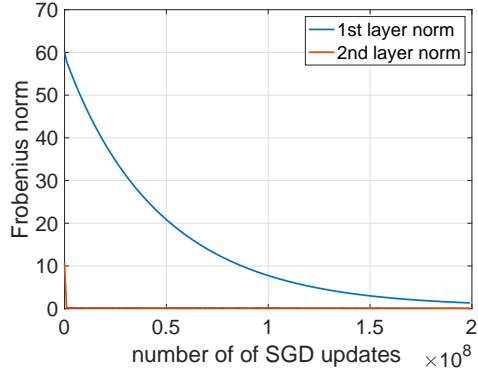} 
\end{center}
\caption{Norm evolution - learning noise without bias neurons with $x\sim N(0,I_d)$. }
\label{fig:weight_decay}
\end{figure}

We trained a network with $4d$  neurons in the hidden layer and observe in Figure \ref{fig:weight_decay} that with no bias neurons the weights decay to zero. With cross entropy, longer training time is needed to nullify the weights compared to the hinge loss (with analogy to items (a) and (b) in Theorem~\ref{thm:v_van}).

\begin{figure}[ht]
\begin{center}
\begin{subfigure}{0.30\textwidth}
\begin{center}
\includegraphics[width=\linewidth]{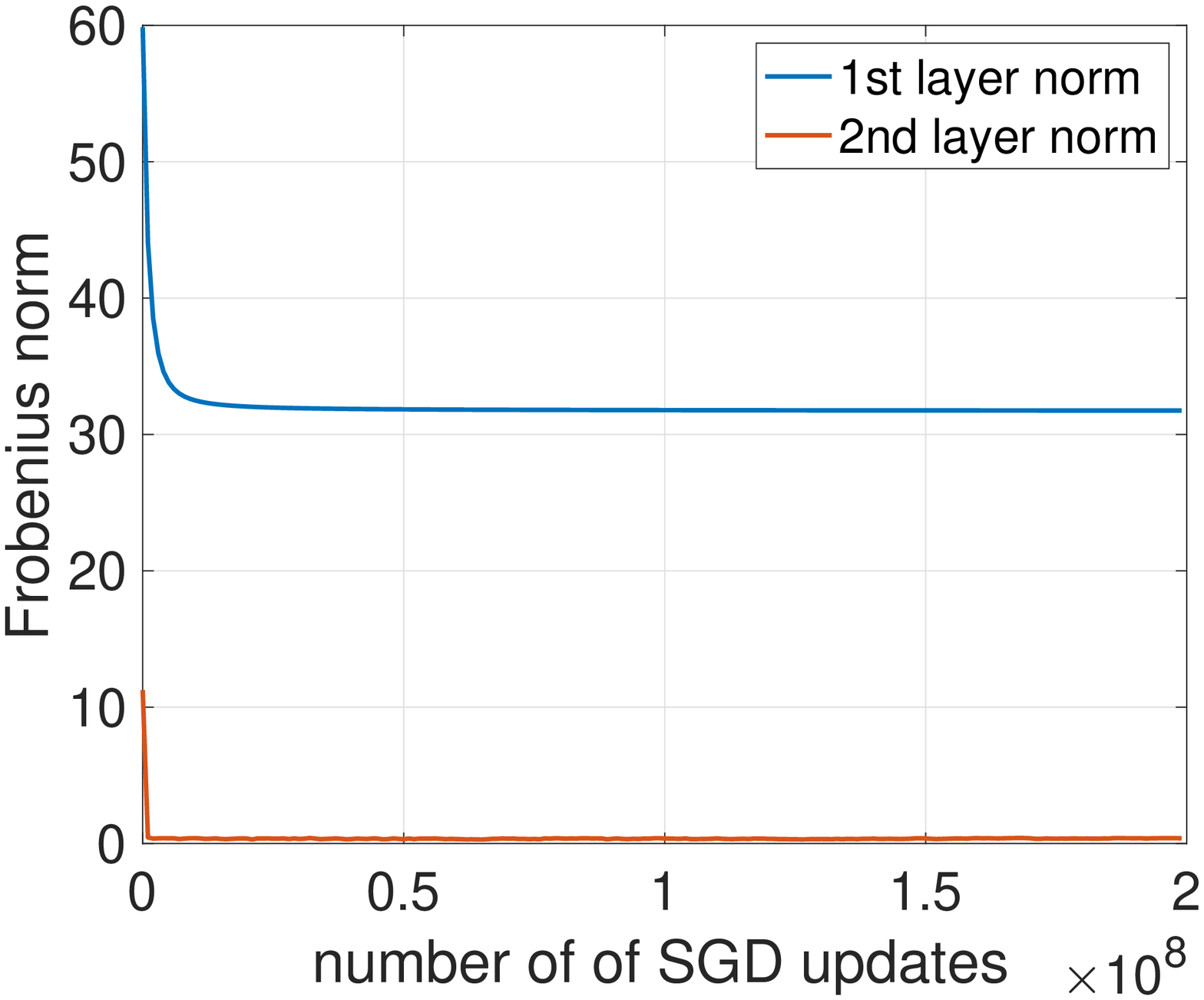} 
\caption{Norm evolution}
\label{fig:bias_weight_decay}
\end{center}
\end{subfigure}
\begin{subfigure}{0.30\textwidth}
\begin{center}
\includegraphics[width=\linewidth]{  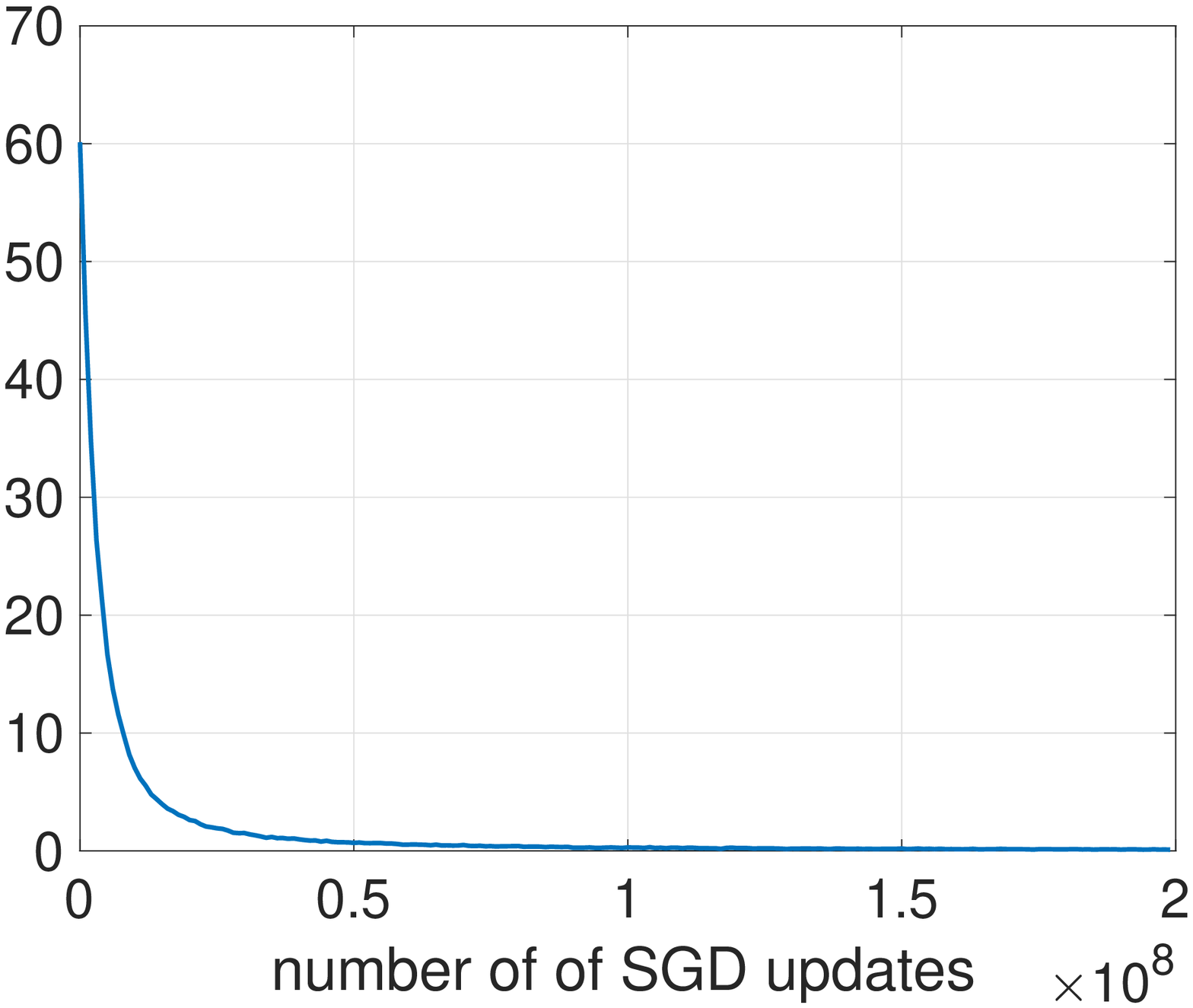}
\end{center}
\caption{Typical number of active neurons}
\label{fig:bias_active}
\end{subfigure}
\begin{subfigure}{0.30\textwidth}
\begin{center}
\includegraphics[width=\linewidth]{  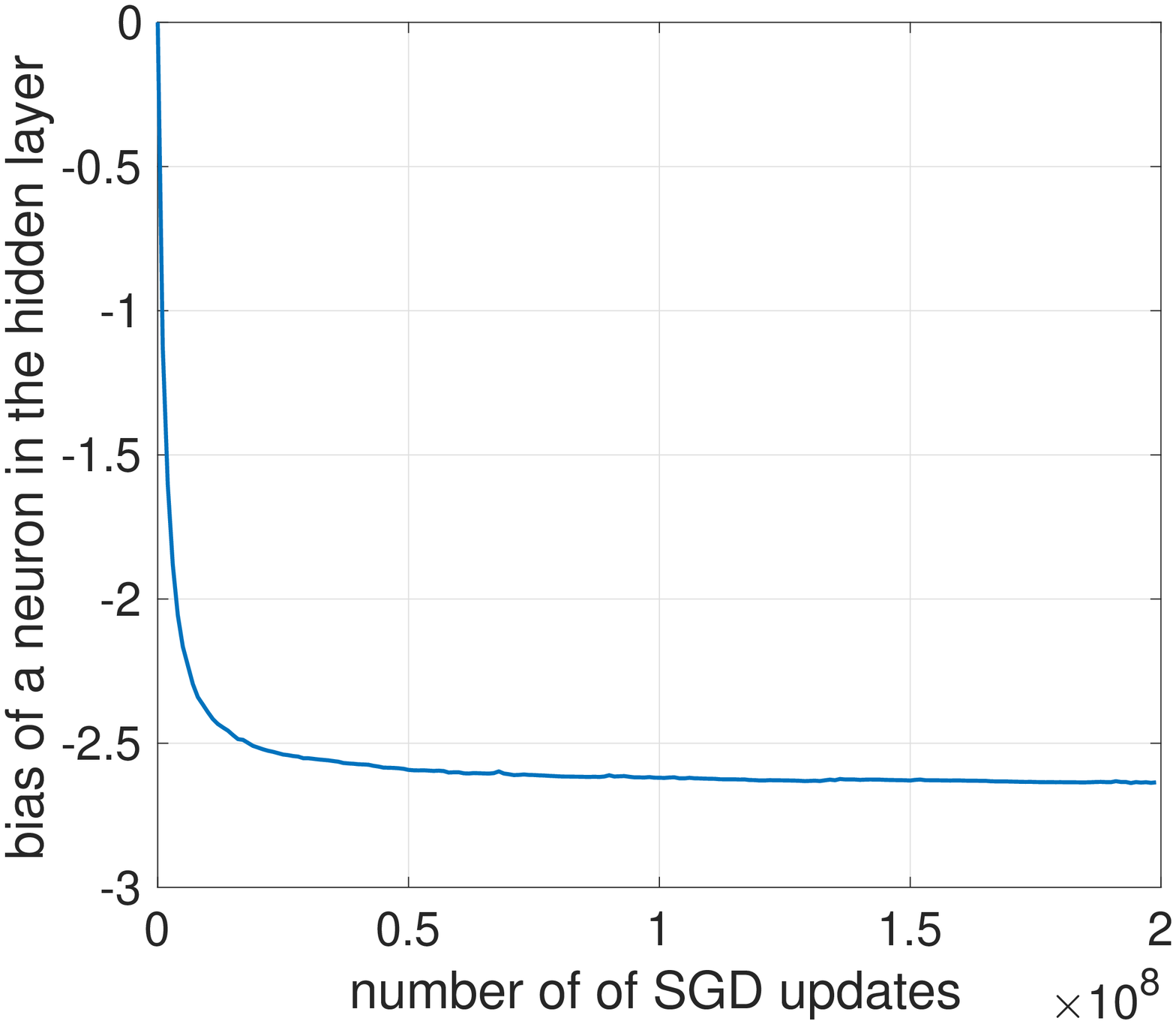}
\end{center}
\caption{Neuron bias evolution}
\label{fig:bias_evo}
\end{subfigure}
 
\caption{Learning noise with bias neurons with $x\sim N(0,I_d)$.}
\label{fig:bias_decay}
\end{center}
\end{figure}

In comparison, Figure~\ref{fig:bias_weight_decay} shows that the weights do not decay to zero when  adding  bias neurons. Figure~\ref{fig:bias_active} shows that adding  bias neurons shuts down the network by making the ReLU neurons stop firing for all inputs. 
The quantity measured in Figure~\ref{fig:bias_active}
is a measure of sparsity of representation
that we use in the subsequent experiments (see Definiton~\ref{def:num_act}).

ReLU neurons dying during training is a well observed phenomenon (for example, see ~\cite{Lu2019DyingRA}), which causes difficulties during the training of neural networks. The experiments in the following sections show that label noise and ReLU death is not necessarily bad for neural networks as our theoretical results for pure label noise might suggest.

\subsection{Random errors sparsify and improve generalization}\label{sec:cube}


In what scenario  adding label noise would make a sizable difference? A reasonable guess would be learning a function that the  network can represent in a sparse way. Such candidate function is the \emph{hypercube boundary function} $\mathcal{F}:C \a \{\pm1\}$, where $C=C_1\cup C_{-1} $, $C_1= \{x\in \RR^d : \lv  x \rv _\infty = 1\}$,  $ C_{-1}   = \{x\in \RR^d : \lv  x \rv _\infty= 1-\epsilon\}$, and $\mathcal{F}(x)=\textbf{1}_{C_1}(x)-\textbf{1}_{C_{-1}}(x)$. This function outputs $1$ if the input comes from the unit hypercube boundary and $-1$ otherwise. 

This function has a simple representation using a one hidden layer neural network $N(x)=W_2\ReLU(W_1\cdot x +B )+b$. $W_1$ is a $2d\times d$ matrix where $W_1(i,i)=1$ and $W_1(i+d,i)=-1$  for $1\leq i\leq d$ and else the entries are 0,
$B$ is a $2d$ vector where $B(i)=-(1-\eps/2)$  for $1\leq i\leq 2d$,  $W_2$ is a $2d$ vector where $W_2(i)=1$  for $1\leq i\leq 2d$ and $b=-0.5$. For this representation, and the input distribution  $\mathcal{D}=\frac{1}{2}U(C_1)+\frac{1}{2}U(C_{-1})$,  the typical number of active neurons
will be  $d\eps$ ($\eps/2$ fraction) for inputs from $C_1$ and no neuron will be active for inputs from $C_{-1}$.

We perform an experiment with a network with $4d$  neurons in the hidden layer.
We randomly pick  a fixed data set according to $D$ with $\eps =0.3$. We investigate the overparametrized regime, so the size of the dataset is $4d$ (the number of parameters is roughly $4d^2$). We explore two training regimes. First, without label noise, second, with label noise
with $p=0.2$. We train both networks with learning rate $h=1/d$ till they reach zero classification error and then some more until the test error stabilizes.



The experiment reveals a significant difference between the networks trained with and without label noise. Figure \ref{fig:sq_active} shows that the network trained with label noise has significantly sparser activation patterns. That is, for a typical input most of the neurons in the hidden layer do not fire. This suggests that neural networks trained with  label noise may generalize better since not all of the network capacity is used.

Indeed, figure \ref{fig:sq_error} shows a clear advantage in the generalization of the network trained with noisy labels. Although it takes this network longer to reach small training error, it makes up for it with a considerable  drop in the test error versus the ``vanilla'' network that  does not perform a lot better than a random guess on unseen data. This behavior is consistent with what was observed by~\cite{blanc2019implicit} in the context of 1D regression (see Figure~2 therein).



\subsection{Noisy labels applied on MNIST and CIFAR-10}

\textbf{MNIST.} We now examine the effect of noisy labels on MNIST. We trained a network with $600$ neurons in the hidden layer (overparametrized network), $10$ neurons in the output layer, and a learning rate  $h=0.01$ which we decrease by a factor of $2$ every $5$ epochs.
This is a multiclass classification task, so we work with the  loss function
\begin{equation}\label{eq:sum_loss}
    \sum_{i=1}^{10} L( -y_i \cdot N(x)_i)
\end{equation}
where the coordinate in $y$ that corresponds to the correct label is $1$, the other entries are $-1$, 
$N(x)_i$ is the output of neuron $i$ in the output layer, and $L(z)=\log (1+\exp(z))$. Our loss function is a sum of $10$ cross entropy losses, each corresponding to one digit. 
This way, we can consider the same properties as discussed throughout the rest of the paper.
For example, we can apply Theorem~\ref{thm:w_decay} to each output neuron separately.
Contrastingly, Theorem~\ref{thm:w_decay} cannot be applied when using the standard cross-entropy loss because the softmax function is applied over all output neurons. And 
indeed, working with the loss in~\eqref{eq:sum_loss} produces sparser representations than working with softmax.

\begin{figure}[h]
 \begin{subfigure}{0.45\textwidth}
\includegraphics[width=0.95\linewidth]{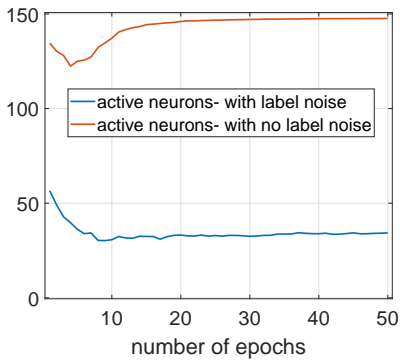}
\caption{Typical number of active neurons}
\label{fig:mnist_active}
\end{subfigure}
\begin{subfigure}{0.45\textwidth}
\includegraphics[width=0.95\linewidth]{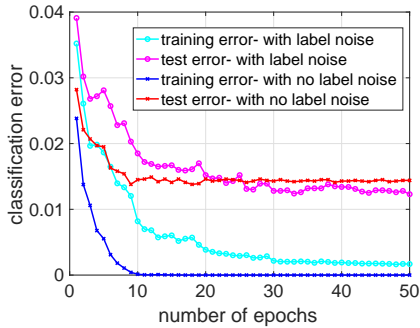} 
\caption{Training and test classification errors}
\label{fig:mnist_error}
\end{subfigure}\hfill
\caption{Learning MNIST with and without noisy labels.}
\label{fig:mnist}
\end{figure}

We trained the network without and with label noise for $p=0.1$. 
Figure~\ref{fig:mnist_error} shows that training
with label noise has better performance in terms of the test
error. This may be explained by the sparser activation patterns (25\% vs. 6\%).

\textbf{CIFAR-10.} To achieve reasonable test accuracy over this dataset ($>90\%$), one must choose a more complex architecture. For CIFAR-10 we chose a standard ResNet-20 with the hyperparameters appearing in~\cite{resnet}. 

To emphasize, in contrast to MNIST, we are working with the standard cross-entropy loss  applied over a softmax of the last layer. This setting is more remote from the theoretical setting we previously explored (convolutional layers, skip connections, batch normalization, etc.). Nevertheless, our findings are consistent with previous experiments and our theoretical discussion. 
Adding noise over the labels sparsifies the firing patterns of the network. 


We trained the network with and without label smoothing with $p=0.1$ (not label noise) to reflect more accurately
what is being done in practice. We achieved test accuracy of slightly more than $92\%$ for both networks. The sparsity was most pronounced in the penultimate layer that consists of 4096 neurons. In this layer, the fraction of active neurons for a typical input for a network trained without label smoothing was $46.9\%$ vs. $27.5\%$ with label smoothing. In general,  deeper layers corresponded to larger sparsity compared to the network trained without label smoothing.

\subsection{Understanding  Label Smoothing Better}

It was observed in~\cite{LS} that a better calibrated network trained with label smoothing  actually achieves better accuracy. Figure 1 of~\cite{LSH} offers some insight on why this happens.  The authors recorded the activations of the penultimate layer of a network trained with and without label smoothing over CIFAR-10. Then, they projected the activations from three classes on a plane. In both cases, three clusters emerged that correspond to the three different labels. With label smoothing the clusters were more tightly packed so the separation between classes is easier and hence the better accuracy. Our work provides some insight to why label smoothing promotes tighter clusters.

We discovered that for MNIST 
learning with label noise or label smoothing generates sparser representations. In these examples
label smoothing actually generates  a sparse representation in the  penultimate layer of a special kind. Most neurons in the penultimate layer can be associated with a specific label. That is, most neurons mostly fire when they are presented with their associated label. More specifically, for every neuron $i$  in the hidden layer (with the corresponding row of weights $W_i$), we counted the number of times it fired for each label $'0','1',...,'9'$. We noticed that these histograms were more concentrated for a network trained with label noise. Examples of such histograms are in the appendix.





This provides an explanation for Figure 1 in~\cite{LSH}.  In this case, picking a plane that intersects three clusters is easy. For every label pick the feature that corresponds the most  with a specific label and then pick a  plane such that the projections of the selected features are large. Since label smoothing generates a stronger association between the features and the labels, the clusters will be tighter.

We remark that although all of our experiments in this section were performed with label noise, except for CIFAR-10, we observe the same qualitative sparsification effects for
label smoothing. Quantitatively, label smoothing generates denser representations compared to label noise.

\section{Conclusion}

Throughout this paper we have studied the effects of misclassification during neural network training. 
Across all the settings we explored, including MNIST and CIFAR-10, label noise or label smoothing induced sparser activation patterns. This even holds over ResNets that are distant from the theoretical setting  we explored, as they incorporate convolutional layers, batch normalization, and so on.  For MNIST, we note that even without any form of label noise, the activations were sparsified (see Figure~\ref{fig:mnist_active}); at initialization, the typical fraction of active neurons is $50\%$, as one would expect, and at the end of training the fraction decreases to $25\%$. This raises the question, does SGD have some implicit bias towards solutions with sparse activation patterns?

\bibliography{bibliography}

\begin{thebibliography}{29}
\providecommand{\natexlab}[1]{#1}
\providecommand{\url}[1]{\texttt{#1}}
\expandafter\ifx\csname urlstyle\endcsname\relax
  \providecommand{\doi}[1]{doi: #1}\else
  \providecommand{\doi}{doi: \begingroup \urlstyle{rm}\Url}\fi

\bibitem[{Szegedy} et~al.(2016){Szegedy}, {Vanhoucke}, {Ioffe}, {Shlens}, and
  {Wojna}]{LS}
C.~{Szegedy}, V.~{Vanhoucke}, S.~{Ioffe}, J.~{Shlens}, and Z.~{Wojna}.
\newblock Rethinking the inception architecture for computer vision.
\newblock In \emph{2016 IEEE Conference on Computer Vision and Pattern
  Recognition (CVPR)}, pages 2818--2826, 2016.

\bibitem[Blanc et~al.(2019)Blanc, Gupta, Valiant, and
  Valiant]{blanc2019implicit}
Guy Blanc, Neha Gupta, Gregory Valiant, and Paul Valiant.
\newblock Implicit regularization for deep neural networks driven by an
  {O}rnstein-{U}hlenbeck like process.
\newblock arXiv:1904.09080, 2019.

\bibitem[M\"{u}ller et~al.(2019)M\"{u}ller, Kornblith, and Hinton]{LSH}
Rafael M\"{u}ller, Simon Kornblith, and Geoffrey~E Hinton.
\newblock When does label smoothing help?
\newblock In H.~Wallach, H.~Larochelle, A.~Beygelzimer, F.~d\textquotesingle
  Alch\'{e}-Buc, E.~Fox, and R.~Garnett, editors, \emph{Advances in Neural
  Information Processing Systems 32}, pages 4694--4703. Curran Associates,
  Inc., 2019.
\newblock URL
  \url{http://papers.nips.cc/paper/8717-when-does-label-smoothing-help.pdf}.

\bibitem[Goodfellow et~al.(2016)Goodfellow, Bengio, and Courville]{dlbook}
Ian Goodfellow, Yoshua Bengio, and Aaron Courville.
\newblock \emph{Deep Learning}.
\newblock MIT Press, 2016.
\newblock \url{http://www.deeplearningbook.org}.

\bibitem[Van~Laarhoven(2017)]{van2017l2}
Twan Van~Laarhoven.
\newblock L2 regularization versus batch and weight normalization.
\newblock arXiv:1706.05350, 2017.

\bibitem[Hanson(1990)]{Hanson1990ASV}
Stephen~Jos{\'e} Hanson.
\newblock A stochastic version of the delta rule.
\newblock \emph{Physica D: Nonlinear Phenomena}, 42\penalty0 (1--3):\penalty0
  265--272, 1990.

\bibitem[Clay and Sequin(1992)]{Clay1992FaultTT}
R.~D. Clay and C.~Sequin.
\newblock Fault tolerance training improves generalization and robustness.
\newblock \emph{[Proceedings 1992] IJCNN International Joint Conference on
  Neural Networks}, 1:\penalty0 769--774 vol.1, 1992.

\bibitem[Murray and Edwards(1994)]{Murray1994EnhancedMP}
Alan~F. Murray and Peter~J. Edwards.
\newblock Enhanced mlp performance and fault tolerance resulting from synaptic
  weight noise during training.
\newblock \emph{IEEE transactions on neural networks}, 5 5:\penalty0 792--802,
  1994.

\bibitem[An(1996)]{an1996effects}
Guozhong An.
\newblock The effects of adding noise during backpropagation training on a
  generalization performance.
\newblock \emph{Neural Computation}, 8\penalty0 (3):\penalty0 643--674, 1996.

\bibitem[Breiman(2000)]{breiman2000randomizing}
Leo Breiman.
\newblock Randomizing outputs to increase prediction accuracy.
\newblock \emph{Machine Learning}, 40\penalty0 (3):\penalty0 229--242, 2000.

\bibitem[Rifai et~al.(2011)Rifai, Glorot, Bengio, and
  Vincent]{Rifai2011AddingNT}
Salah Rifai, Xavier Glorot, Yoshua Bengio, and Pascal Vincent.
\newblock Adding noise to the input of a model trained with a regularized
  objective.
\newblock \emph{ArXiv}, abs/1104.3250, 2011.

\bibitem[Sukhbaatar and Fergus(2014)]{sukhbaatar2014learning}
Sainbayar Sukhbaatar and Rob Fergus.
\newblock Learning from noisy labels with deep neural networks.
\newblock arXiv:1406.2080, 2014.

\bibitem[Maennel et~al.(2020)Maennel, Alabdulmohsin, Tolstikhin, Baldock,
  Bousquet, Gelly, and Keysers]{maennel2020neural}
Hartmut Maennel, Ibrahim Alabdulmohsin, Ilya Tolstikhin, Robert~JN Baldock,
  Olivier Bousquet, Sylvain Gelly, and Daniel Keysers.
\newblock What do neural networks learn when trained with random labels?
\newblock 2020.

\bibitem[Abbe and Sandon(2020)]{Abbe2020PolytimeUA}
Emmanuel Abbe and Colin Sandon.
\newblock Poly-time universality and limitations of deep learning.
\newblock \emph{ArXiv}, abs/2001.02992, 2020.

\bibitem[Lu et~al.(2019)Lu, Shin, Su, and Karniadakis]{Lu2019DyingRA}
Lu~Lu, Yeonjong Shin, Yanhui Su, and George~Em Karniadakis.
\newblock Dying relu and initialization: Theory and numerical examples.
\newblock \emph{ArXiv}, abs/1903.06733, 2019.

\bibitem[Arnekvist et~al.(2020)Arnekvist, Carvalho, Kragic, and
  Stork]{Arnekvist2020TheEO}
Isac Arnekvist, Jo{\~a}o~Frederico Carvalho, Danica Kragic, and
  Johannes~Andreas Stork.
\newblock The effect of target normalization and momentum on dying relu.
\newblock \emph{ArXiv}, abs/2005.06195, 2020.

\bibitem[Douglas and Yu(2018)]{Douglas2018WhyRU}
Scott~C. Douglas and Jiutian Yu.
\newblock Why relu units sometimes die: Analysis of single-unit error
  backpropagation in neural networks.
\newblock \emph{2018 52nd Asilomar Conference on Signals, Systems, and
  Computers}, pages 864--868, 2018.

\bibitem[{Mehta} et~al.(2019){Mehta}, {Kim}, and {Theobalt}]{sparse1}
D.~{Mehta}, K.~I. {Kim}, and C.~{Theobalt}.
\newblock On implicit filter level sparsity in convolutional neural networks.
\newblock In \emph{2019 IEEE/CVF Conference on Computer Vision and Pattern
  Recognition (CVPR)}, pages 520--528, 2019.
\newblock \doi{10.1109/CVPR.2019.00061}.

\bibitem[Du et~al.(2018)Du, Hu, and Lee]{Du2018AlgorithmicRI}
Simon~S. Du, Wei Hu, and Jason~D. Lee.
\newblock Algorithmic regularization in learning deep homogeneous models:
  Layers are automatically balanced.
\newblock \emph{ArXiv}, abs/1806.00900, 2018.

\bibitem[Hanin(2018)]{hanin2018neural}
Boris Hanin.
\newblock Which neural net architectures give rise to exploding and vanishing
  gradients?
\newblock In \emph{Advances in Neural Information Processing Systems}, pages
  582--591, 2018.

\bibitem[Neyshabur et~al.(2015)Neyshabur, Tomioka, and
  Srebro]{Neyshabur2015InSO}
Behnam Neyshabur, Ryota Tomioka, and Nathan Srebro.
\newblock In search of the real inductive bias: On the role of implicit
  regularization in deep learning.
\newblock \emph{CoRR}, abs/1412.6614, 2015.

\bibitem[Gunasekar et~al.(2018)Gunasekar, Lee, Soudry, and
  Srebro]{Gunasekar2018ImplicitBO}
Suriya Gunasekar, Jason~D. Lee, Daniel Soudry, and Nathan Srebro.
\newblock Implicit bias of gradient descent on linear convolutional networks.
\newblock In \emph{NeurIPS}, 2018.

\bibitem[Soudry et~al.(2018)Soudry, Hoffer, Gunasekar, and
  Srebro]{Soudry2018TheIB}
Daniel Soudry, Elad Hoffer, Suriya Gunasekar, and Nathan Srebro.
\newblock The implicit bias of gradient descent on separable data.
\newblock \emph{J. Mach. Learn. Res.}, 19:\penalty0 70:1--70:57, 2018.

\bibitem[Soudry et~al.(2017)Soudry, Hoffer, Nacson, Gunasekar, and
  Srebro]{soudry2017implicit}
Daniel Soudry, Elad Hoffer, Mor~Shpigel Nacson, Suriya Gunasekar, and Nathan
  Srebro.
\newblock The implicit bias of gradient descent on separable data, 2017.

\bibitem[Brutzkus et~al.(2018)Brutzkus, Globerson, Malach, and
  Shalev-Shwartz]{SGD}
Alon Brutzkus, A.~Globerson, Eran Malach, and S.~Shalev-Shwartz.
\newblock Sgd learns over-parameterized networks that provably generalize on
  linearly separable data.
\newblock \emph{ArXiv}, abs/1710.10174, 2018.

\bibitem[Idelbayev(2020)]{resnet}
Yerlan Idelbayev.
\newblock Reproducing cifar10 experiment in the resnet paper, 2020.
\newblock URL
  \url{https://colab.research.google.com/github/seyrankhademi/ResNet_CIFAR10/blob/master/CIFAR10_ResNet.ipynb}.

\bibitem[Johnstone(2001)]{chi}
Iain~M. Johnstone.
\newblock Chi-square oracle inequalities.
\newblock \emph{Lecture Notes-Monograph Series}, 36:\penalty0 399--418, 2001.
\newblock ISSN 07492170.
\newblock URL \url{http://www.jstor.org/stable/4356123}.

\bibitem[Shamir(2011)]{azuma}
Ohad Shamir.
\newblock A variant of {A}zuma's inequality for martingales with subgaussian
  tails.
\newblock \emph{CoRR}, abs/1110.2392, 2011.
\newblock URL \url{http://arxiv.org/abs/1110.2392}.

\bibitem[Tao(2015)]{anti}
Terence Tao.
\newblock Variants of the central limit theorem, 2015.
\newblock URL
  \url{https://terrytao.wordpress.com/2015/11/19/275a-notes-5-variants-of-the-central-limit-theorem/#more-8566}.

\end{thebibliography}

\newpage

\appendix

\section{Continuity of $a(p)$}\label{thm:rig_con}

Let $A_N(x)$ be the number of active neurons for an input $x$ of a ReLU neural nertwork with one hidden layer $N$ trained in the presence of label noise with noise parameter $p$ over a dataset $S=\{(x_1, y_1),(x_2, y_2),...,(x_n, y_n)\}$. Then, $A_{N}(x)$ depends on several hyperparameters, specifically: the weights initialization $W\uz$,  the learning rate $h$, the number of training  iterations $T$, the ordering of the training sequence $ord=i_1,...,i_T$ where $1\leq i_t\leq n$, and the noise parameter $p$. 
These hyperparameters induce the training sequence   $S_p=(x_{i_1},c_1 \cdot y_{i_1}),(x_{i_2},c_2 \cdot y_{i_2}),...,(x_{i_T},c_T \cdot y_{i_T})$ where $c_i=2b_i-1$ and $b_i$ are   i.i.d. Bernoulli  random variables with parameter $1-p$. Note that the final network $N$ is completely determined by $S_p$.

 
For a fixed dataset $S$ and fixed $h$ and $T$,  we define $a(p)$ for $p \in [0,1] $ as the the typical number of active neurons averaged over all randomness of SGD.
\begin{align*}
    a(p):&= \EE_{S_p} \EE_{x \sim U\{x_1,...,x_n\} }  A_{N}(x).
\end{align*}

For any $p$, we can write
\begin{align*}
    a(p) &= \EE_{S_p} \EE_{x \sim U\{x_1,...,x_n\} }  A_{N}(x) 
        =\EE_{ (b_1,...b_T)} \EE_{ord}  \EE_{x \sim U\{x_1,...,x_n\} }  A_{N}(x)   \\            \\
    &=\EE_{ (b_1,...b_T)} \left[\EE_{ord}  \EE_{x \sim U\{x_1,...,x_n\} }  A_{N}(x) ~ | ~ (b_1,...b_T) \right] \\
    &= \sum_{i=0} ^T  p^{T-i} (1-p)^{i} \sum_{ \sum _{t=0}^T b_t=i} \left[ \EE_{ord} \EE_{x \sim U\{x_1,...,x_n\} }  A_{N}(x) ~ | ~ (b_1,...b_T)\right] 
\end{align*}


 Clearly, $a(p)$ is a polynomial in $p$ and thus continuous in $[0,1]$.

\section{Special cases of Theorem 
\ref{thm:w_decay}}

Before the general proof, we provide proofs for two special cases,  a single neuron and a network with one hidden layer.  This provides specific insight for these cases as we explore them in more detail in the paper.  

\begin{proof}[Proof for a single neuron] 
Let $N(x)=V \cdot x$. 
Because $y V \tu \cdot x < 0$,
the derivative of the loss is $L'(-y V\tu \cdot x)=1$. 
So, $V \tup =V\tu + hyx$ 
and 
\begin{align*}
\lv V \tup  \rv ^2&=(V\tu + hyx)\cdot (V\tu + hyx)
\\&=\lv V\tu \rv ^2 +2 hy V \tu \cdot x +h^2 \lv x \rv ^2\;.
\end{align*}
By assumption $y V \tu \cdot x < 0$. 
Thus, if $h$ is small enough then 
the norm of $V$ decreases.
\end{proof}

\begin{proof}[Proof for a network with one hidden layer] Let $N(x)=V\cdot  \R (W\cdot x)$. 
As for a single neuron,
$V \tup =V\tu + hy\R (W\tu \cdot x)$ and 
\begin{align*}
    \lv V \tup  \rv ^2 
    =\lv V\tu \rv ^2 +2 hyN\tu (x)+h^2 \lv \R (W\tu \cdot x) \rv ^2.
\end{align*}
The Frobenius norm of $V$ again decreases.

To investigate the norm of $W\tu$, let us consider the $i$-th row $W\tu _i$. 
By the chain rule/back-propagation,
the derivative is non-zero only if $W\tu _i\cdot x > 0$. In words, for the gradient according to  $W\tu _i$ to be non-zero, the network must misclassify $x$ and the corresponding  neuron $i$ must fire when presented with $x$. 

In this case, if $v_i\tu$ is the $i$-th coordinate of $V\tu$, we have $W\tup _i= W\tu _i +hyv_i\tu x$ and  
\begin{align*}
\lv W\tup _i \rv ^2
&= (W\tu _i +hyv_i\tu  x) \cdot (W\tu _i +hyv_i\tu  x)
\\&=
\lv W\tu _i \rv ^2+ 2hyv_i \tu W\tu _i  \cdot x  + h^2(v_i\tu )^2\lv x\rv^2
\;.
\end{align*}
By assumption $W\tu _i  \cdot x>0$, so $ \lv W\tup _i \rv ^2= 
\lv W\tu _i \rv ^2+ 2hyv_i \tu \R (W\tu _i  \cdot x  )+ h^2(v_i\tu )^2\lv x\rv^2  $.
We cannot deduce if the norm of $W_i$ decreases or increases, however, summing over all $i$ (including non-firing neurons) yields 
\begin{align*}
&\lv W\tup \rv ^2
 = \sum_i  \lv W\tup _i \rv ^2 
\\ &\quad\leq  
\sum _i \lp \lv W\tu _i \rv ^2+ 2hyv_i \tu \R (W\tu _i  \cdot x  )+ h^2(v_i\tu )^2\lv x\rv^2 \rp  
\\&\quad
 = 
\lv W \tu \rv ^2+ 2hyV \tu \cdot \R (W\tu   \cdot x  )+ h^2\lv V\tu \rv ^2\lv x\rv^2 
\\&\quad
= \lv W \tu \rv ^2+ 2hyN(x)+ h^2\lv V\tu \rv ^2\lv x\rv^2   . 
\end{align*}

The inequality follows since the non-firing neurons do not contribute anything to the middle term in the sum and the last term in the sum is non-negative. We see that although the norm of some $W_i\tu$ might increase, the norm of $W\tu$ must decrease for a small enough learning rate.
\end{proof}

\section{Proof of Theorem~\ref{thm:w_decay}}\label{pf:w_decay}
We prove the statement for networks without bias terms.

For the weights in the last layer, the proof is modeled after
our discussion of a single neuron case. Let $V$ be the weights
in the last layer and $a$ denote the activations of the
last layer of neurons. Recall that
we are considering a sample $(x,y)$ where $yN(x)=y(a\cdot V)<0$
and let $\ell':=L'(yN(x))<0$.
By backpropagation, we have
\begin{align*}
    \frac{\partial L(yN(x))}{\partial V_i}
    =\ell' y a_i
\end{align*}
and accordingly the loss gradient
with respect to $V$ is 
$\frac{\partial L(yN(x))}{\partial V}=\ell' ya$. At the same time,
\begin{align*}
    \|V^{(t+1)}\|^2
    &=
    \big(V^{(t)}-h\ell' ya\big)\cdot
    \big(V^{(t)}-h\ell' ya\big)
    =\|V^{(t)}\|^2-2h\ell'y(a\cdot V)+O(h^2)
    <\|V^{(t)}\|^2\;,
\end{align*}
since $h\ell'y(a\cdot V)>0$.

For the preceding layers, let $W$ and $V$ denote
two subsequent layers in the network. By the same
calculation, 
\begin{align}
    \|W^{(t+1)}\|^2
    &=\left(W^{(t)}-h\frac{\partial L(yN(x))}
    {\partial W^{(t)}}\right)\cdot
    \left(W^{(t)}-h\frac{\partial L(yN(x))}
    {\partial W^{(t)}}\right)
    \nonumber\\
    &=\|W^{(t)}\|^2 -2h\left(W^{(t)}\cdot 
    \frac{\partial L(yN(x))}{\partial W^{(t)}}\right)+O(h^2)\;.
    \label{eq:01}
\end{align}
On the other hand, consider the gradient flow algorithm
with continuous time parameter $t$ run on our neural network and one sample
$(x,y)$. In this gradient flow we have
\begin{align}\label{eq:02}
    \frac{\mathrm{d}\| W\|^2}{\mathrm{d} t}
    =2\left(W\cdot \frac{\partial L(yN(x))}{\partial W}\right)\;.
\end{align}
However, we can now invoke Corollary~2.1 and Theorem~2.3 of~\cite{Du2018AlgorithmicRI} that state
that for all network architectures
$\frac{\mathrm{d}\|W\|^2}{\mathrm{d}t}=\frac{\mathrm{d}\|V\|^2}{\mathrm{d}t}$. Applying this
together with~\eqref{eq:01} and~\eqref{eq:02} for $V$ and $W$ and induction we get
$\|W^{(t+1)}\|^2<\|W^{(t)}\|^2$ for small enough $h$.\qed
\newpage
\section{Proof of Theorem~\ref{thm:v_van}}

 
 

\begin{proof}
Fix an initialization $V^{(0)}$. 

Assume (a) $L''(0)=m>0$. 
This means that for some neighborhood of $h=0$ it holds 
$m_2:=L'(d\sqrt{h}) > L'(0) + \frac{dm \sqrt{h}}{2}  $ 
and $m_1:= L'(-d\sqrt{h}) < L'(0) - \frac{dm\sqrt{h}}{2}  $. We prove the norm of $\lv V\tu  \rv $ will reach below $N= 100 \max \{ d\sqrt{h} M^2/m ,d\sqrt{h} \}    $.



Let $(x_1,y_1),(x_2,y_2),...$ be the random instances presented during the optimization. The weights change  according to 
$V^{(t)}=V^{(t-1)}+hL' y_t 
x_t$, where we write $L'=L'(-y_tV^{(t-1)}\cdot x_t)$ for short,
and hence the norm  $\lv V^{(t)}\rv^2
=\lv V^{(t-1)}\rv^2+2hL'y_tV^{(t-1)}\cdot x_t
+\big(hL'\lv x_t\rv\big)^2$.

We denote by $c_t$ and $r_t$  the random variables such that 
$L'y_t V \tum \cdot x_t=c_t \lv V \tum \rv $ and $\sqrt{h} M^2  \lv x_t\rv^2  =0.01 mr_t \lv V \tum \rv $.

On the one hand,


$\lv V\tu \rv ^2 -\lv V\tum \rv ^2= 2hL'y_t(V\tum \cdot x_t) + (hL'\lv x_t\rv)^2 < 2hc_t\lv V\tum \rv  +0.01   h^{3/2} mr_t \lv V\tum \rv          $.


On the other hand, since $(L-R)^2\ge 0$ is equivalent to
$L^2-R^2\ge 2LR-2R^2=2(L-R)R$, we have


$\lv V\tu \rv ^2 -\lv V\tum \rv ^2
\geq 2 (\lv V\tu \rv -\lv V\tum \rv) \lv V\tum \rv$.

Combining the above inequalities yields 
$   \lv V\tu \rv -\lv V\tum \rv <      hc_t       + 0.005h^{3/2}m r_t       $.

We will now show that, conditioned on $\lv V\tum \rv$, we have
$\EE c_t  < -md\sqrt{h}/4$ and
$\EE r_t  <  1 $. In other words, the evolution of the norm is bounded
by a negatively biased random walk, with the expectation of the order
$-md\sqrt{h}$.

The bound on $\EE r_t$ holds because for each $t$ we assume $ \lv V\tum \rv > 100  d\sqrt{h} M^2/m  $ (otherwise we are done), so
$\EE r_t=\frac{100\sqrt{h}M^2}{m\lv V\tum\rv}\EE\lv x_t\rv^2
<\frac{1}{d}\EE\lv x_t\rv^2=1$.

For the bound on $\EE c_t$, consider that the distribution of $p_{t}=y_t\frac{V\tum}{\lv V\tum\rv} 
\cdot x_{t} \sim N(0, 1) $, therefore
$ P( p_t > 0.01) = P( p_t < -0.01) > 0.49$.
Furthermore, recall that
$c_t=L'(-\lv V\tum\rv p_t)\cdot p_t$. Since $L'(\cdot)$ is increasing,
by symmetry considerations $\EE \Big[c_t\cdot\mathbbm{1}(|p_t|\leq 0.01)\Big]< 0$.
On the other hand, since we assume  $\lv V\tum \rv > 100d\sqrt{h}$
we have

$\EE\Big[c_t \cdot \mathbbm{1}(p_t > 0.01)\Big]=
\EE\Big[L'(-\lv V\tum\rv p_t)\cdot p_t\cdot\mathbbm{1}(p_t>0.01)\Big]
\leq m_1\EE\Big[p_t\cdot\mathbbm{1}(p_t>0.01)\Big]$, 
and similarly $\EE\Big[c_t\cdot \mathbbm{1}(p_t < -0.01)\Big]\le 
m_2\EE\Big[p_t\cdot\mathbbm{1}(p_t<-0.01)\Big]$.
Putting these inequalities together indeed gives

$\EE[c_t]<-(m_2-m_1)\EE\Big[p_t\cdot\mathbbm{1}(p_t>0.01)\Big]
<-(m_2-m_1)/4\le-dm\sqrt{h}/4$.

To complete the proof, we show concentration 
for our two random variables $r_t$ and $c_t$. 

Since $0\le r_t<\frac{1}{d}\lv x_t\rv^2$ and
random variables $\lv x_t\rv^2$ are iid chi-squared
with $d$ degrees of freedom,
we can apply a standard tail bound for chi-squared distribution (see Lemma 6.1 in~\cite{chi}):

$$ P( \sum_{t=1}^T r_t - T >  T ) < 
P( \sum_{t=1}^T \lv x_t\rv^2 - dT >  dT ) 
< \exp(-\frac{dT}{8} )\;.
$$


For the concentration of $c_t$, note that $c_t+md\sqrt{h}/4$ is a supermartingale
with respect to $V\tu$. Furthermore, $c_t\mid V\tum$ is sub-Gaussian 
(because $p_t\mid V\tum $ is a Gaussian). Therefore, we can
apply the Azuma concentration inequality for sub-Gaussian differences
in~\cite{azuma}. Specifically, conditioned on $V\tum$,
for any $a>0$ we have

$$\Pr(c_t>a)=\Pr\big(L'(-\lv V\tum\rv p_t)\cdot p_t>a\big)<
\Pr(p_t>a/M)<\exp(-a^2/2M^2)$$

and similarly $\Pr(c_t<-a)<\exp(-a^2/2M^2)$. Hence, we can apply Theorem~2 
from~\cite{azuma} and get

$$ P\Big( \sum_{i=1}^T c_i > -md\sqrt{h}T/4 + \epsilon T \Big) < 
\exp \left(-\frac{T\epsilon ^2}{64M^2}\right)\;. $$

We choose $T=\lceil \frac{7\lv V^{(0)}\rv}{mdh^{3/2}} \rceil$ and $\epsilon = M/T^{0.4}$. 
Except with  
probability $\exp(-\Omega(T^{0.2}))$, 
for a small enough learning rate, we will have that 
$$\lv V^{(T)} \rv < \lv V^{(0)}  \rv - \frac{3\lv V^{(0)}  \rv}{2} + M(\frac{7\lv V^{(0)}  \rv}{mdh^{3/2}})^{0.6}h +0.01 \frac{7\lv V^{(0)}\rv}{d} < 
-0.4\lv V^{(0)}\rv +O(h^{0.1}), $$ which would be a contradiction,
meaning that the norm of $V\tu$ must have
fallen below $N$ before time $T$.

\medskip

For (b), consider $m_2:=L'(dh)$, $m_1:=L'(-dh)$, and 
$N=100 \max\{\frac{dhM^2}{m} , dh \}$, where  $m<m_2-m_1$ by  
definition. The proof now follows in similar lines, only in this 
case,  $T=\lceil \frac{3\lv V^{(0)}\rv}{mh} \rceil$.
\end{proof}



\begin{proof}
By orthogonality of $e_1,\ldots,e_d$, any update in the network when presented with $e_i$ does not affect the activations corresponding to $e_j$ for $j\neq i$. To see this, let $W\tu _r$ be the $r$-th row of $W$ at time $t$. Then, $W\tup _r=W\tu _r\pm he_i$, so $W\tup _r \cdot e_j =(W\tu _r \pm he_i) \cdot e_j= W\tu _r\cdot e_j$. 

This means we can focus only on the first
column of $W=W\tu$ and fix $x=e_1$. 
At time $t$, let 
$Pos^{(t)}=\{1\le i\le k: W_{i1}>0\}$ and 
$Neg^{(t)}=\{k+1\le i\le 2k: W_{i1}>0\}$ be the sets of ``live'' weights. Note that we have
\[
N^{(t)}(x)=\sum_{i\in Pos^{(t)}} W^{(t)}_{i1}-\sum_{i\in Neg^{(t)}} W^{(t)}_{i1}\;.
\]
Observe that $Pos^{(t+1)}\subseteq Pos^{(t)}$ and 
$Neg^{(t+1)}\subseteq Neg^{(t)}$. Therefore, after a certain amount of time
these sets must stabilize, i.e., $Pos^{(t+1)}=Pos^{(t)}$ always.
Consider a time $t$ after this point.

Assume that $N^{(t)}>0$, the opposite case being similar.
In that case, by the formulas for the gradient,
the update happens only when $y=-1$ and it decreases
$W_{i1}$ by $h$ if $i\in Pos^{(t)}$ and increases $W_{i1}$ by $h$
if $i\in Neg^{(t)}$, leaving all other entries of $W^{(t)}$ unchanged.
Let $\Delta:=h\cdot\big(|Pos|+|Neg|\big)$. Note that $\Delta\ge 0$ and
that $N^{(t+1)}=N^{(t)}-\Delta$.

Now, if $N^{(t)}\ge \Delta$, then the output of the network decreases exactly by $\Delta$.
On the other hand,
if $N^{(t)}<\Delta$, then also $|N^{(t+1)}|<\Delta$.
Therefore, the value of $|N^{(t)}|$ must decrease below $\Delta$ in
a finite number of steps and stay there forever.

Since $|\Delta|\le 2kh$, the proof is concluded.
\end{proof}

\section{Proof of Theorem~\ref{thm:step_size}}


\begin{proof}
As in the proof of Theorem \ref{thm:network_relu_nobias}, we can only focus on $x=e_1$.

\paragraph{Learning rate $h\geq1$} During training, after encountering $x$ three times with misclassified labels we will have  $\ReLU(We_1)=\overrightarrow{0}$. To see this, assume w.l.o.g. that at initialization $N^{(0)} (x)>0$. When encountering $(x, -1)$ for the first time at time $t$,  all the weights associated with  elements in $Pos^{(t)}$ will become negative because they lie in $[0,1]$ and the learning rate $h\geq1$, so $Pos\tup = \emptyset$. Also, the weights associated with $Neg\tup$ will now have values in $[1,2]$. 

Since $\{Pos^{(i)}\}_{i=1}^{\infty}$ is a nested sequence, $Pos^{(i)}$ will remain empty for $i>t$. This means that updates now will occur only when encountering $(x,1)$. After two such encounters $Neg^{(i)}$ will be the empty set and in total $\ReLU(Wx)=\overrightarrow{0}$.

\paragraph{Learning rate $h\leq 1/k$} By Hoeffding's inequality, for $X_1,...,X_n\in [0,1]$ i.i.d. random variables it holds    $P(| \sum_{i=1}^k  X_i-n\EE X_1  |> n\epsilon ) < 2\exp(-2n\epsilon ^2)$. Then, with high probability (with respect to $k$), all items hold:
\begin{enumerate}
    \item
    $k/2+k^{0.6}>|Pos^{(0)}|>k/2-k^{0.6}$. Since the probability of a weight to be included  in $Pos^{(0)}$ is $1/2$, we have $X_i\sim\text{Bernoulli}(1/2)$, $n=k$, and $\epsilon= \frac{1}{k^{0.4}}$. 
    \item $k/2+k^{0.6}>|Neg^{(0)}|>k/2-   k^{0.6}$. Similarly, as above.
    \item $3 <|N^{(0)}(x)|< \frac{1}{2} k^{0.6}$. $N^{(0)}(x)=\sum_{i=1}^k X_i-\sum_{i=k+1}^{2k} X_i$ where  $X_i$ are i.i.d. random variables such that with probability $1/2$ we have $X_i=0$ and with probability $1/2$ we have $X_i$ is uniform in $[0,1]$.  For the upper bound, we apply Hoeffding's inequality for each of the sums with $\epsilon=\frac{1}{4k^{0.4}}$. For the lower bound (anti-concentration), we apply Theorem 8 in~\cite{anti}.
    \item $|\{i~|~i\in Pos\uz,~ W_{i1} < \frac{1}{k^{0.4}}\}| < 2k^{0.6}$. We upper bound the number of small weights $<\frac{1}{k^{0.4}}$ in $Pos\uz$, so we use $X_i\sim\text{Bernoulli}(\frac{1}{k^{0.4}})$ and $\epsilon=\frac{1}{k^{0.4}}$.
    \item $|\{i~|~i\in Neg\uz,~ W_{i1} < \frac{1}{k^{0.4}}\}| < 2k^{0.6}$.  Similarly, as above.
    
\end{enumerate}

Assume $N^{(0)}(x)>0$, the opposite case being similar. As long as  $N^{(t)}(x)>0$, updates only occur  when encountering $(x,-1)$. In these cases, $N\tup(x)=N\tu(x)-h(|Pos\tu|+|Neg\tu|)$.
In these updates, $Neg\tu$ does not change.
Furthermore, since at each step,
each weight in $Pos\tu$ decreases by $h$,
then for $t<\frac{1}{hk^{0.4}}$ the total decrease
is at most $th<1/k^{0.4}$, and by item~4
we have $|Pos\tu|>k/2-3k^{0.6}$.
But then, by item~3, we have
$N\tu(x)<k^{0.6}/2-th(k/2-3k^{0.6}+k/2-k^{0.6})$
and accordingly $N\tu<0$ for some
$t<\frac{1}{hk^{0.4}}$.

From now on, let $t$ be the first time
with $N\tu<0$. By items~1, 2, and 3, it holds that
$N\tu(x)>3-(k+2k^{0.6})ht$. Since $h\le 1/k$, we have
 $t\ge 2$. By the upper bounds of items~1 and 2, we have $N\tu>-(k+2k^{0.6})h$  

Now, since 
$N\tup=N\tu+h(|Pos\tu|+|Neg\tu|)$
and by the lower bounds 
$|Pos\tu|,|Neg\tu|>k/2-3k^{0.6}$ we
have $N\tup-N\tu>0.9kh$. In this reversed
situation, clearly, $Pos\tup=Pos\tu$.
At the same time, also 
$Neg\tup=Neg\tu$ since $t\ge 2$
so each weight from $Neg\tu$ exceeds
$2h$. For another update, we will have   $Neg\tupp=Neg\tup$, $Pos\tupp=Pos\tup$, so $N\tupp-N\tup>0.9kh$ and $N\tupp>0$. Similarly, from time
$t+2$ on the sets $Pos\tu$ and $Neg\tu$ 
are stable and the sequence
$N\tu$ becomes periodic with a period of 2
and alternating signs.

Finally, since $Pos\tu$ and $Neg\tu$  are stable,
 we have $|Pos\tu|,|Neg\tu|> k/2 - 3k^{0.6}$ for all $t$ and the vector
$\ReLU(Wx)$  indeed has at least
$k-o(k)$ non-zero coordinates.
\end{proof}

\section{Label Noise and a Sparse Representation for MNIST}

\begin{figure}[h]
 \begin{subfigure}{0.9\textwidth}
\includegraphics[width=0.90\linewidth]{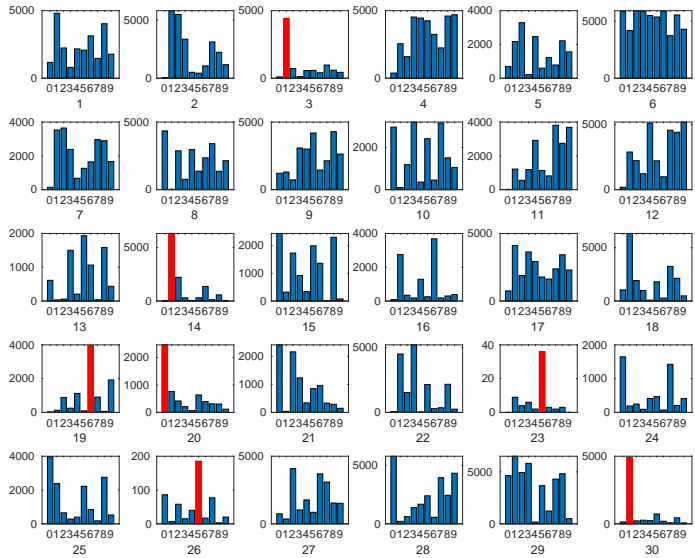}
\caption{No label noise $p=0$}
\label{fig:histp0}
\end{subfigure}
\begin{subfigure}{0.9\textwidth}
\includegraphics[width=0.90\linewidth]{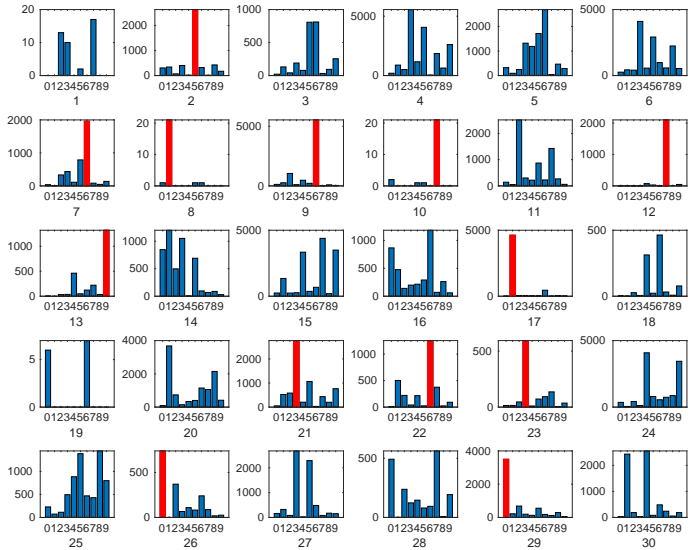} 
\caption{With label noise $p=0.1$}
\label{fig:histp10}
\end{subfigure}
\caption{Learning MNIST with and without noisy labels. Every subplot is a histogram for a random neuron which quantifies how many times the neuron fired for each digit in MNIST. When a neuron is \emph{digit associated}, a red bar highlights the associated digit. }
\label{fig:histo}
\end{figure}

Figure~\ref{fig:histo} demonstrates how label noise induces a special sparse representation for MNIST when training a one hidden layer neural network with $600$ neurons. Figure~\ref{fig:histp0} (without label noise) and Figure~\ref{fig:histp10} (with label noise $p=0.1$) each show histograms for 30 randomly selected neurons from the hidden layer after training over MNIST. Each histogram shows how many times a neuron fired for each digit '0',...,'9'. 

Qualitatively, we can observe more neurons that are associated with a specific digit in Figure~\ref{fig:histp10} compared to Figure~\ref{fig:histp0}. Quantitatively,  we say that a neuron is \textbf{\emph{digit associated}} if it fires at least twice as much for some digit compared to any other digit. For example, in Figure~\ref{fig:histp0}, plots 3, 14, 19, 20, 23, 26, and 30  are associated with the digits '1', '1', '6', '0', '5', '5', and '1', respectively; in Figure~\ref{fig:histp10},  plots 2, 7, 8, 9, 10, 12, 13, 17, 21, 22 ,23, 26, and 29  are associated with the digits '5', '6', '1', '6', '7', '7', '9', '1', '3', '6', '3', '0', and '0', respectively. 

The notion of associated digit is reminiscent to the notion of a grandmother cell in neuroscience; a neuron which fires if and only if one thinks of his grandmother. Using our quantification, training without label noise corresponds to 104 digit associated neurons versus 222 digit associated neurons with label noise $p=0.1$. 

\end{document}